\documentclass[11pt, letterpaper, logo, onecolumn, copyright, numbering]{acc}

\newcommand{\minconf}{\operatorname{minconf}}

\usepackage{url}

\usepackage{natbib}

\usepackage{amsthm}

\usepackage{times}

\definecolor{darkblue}{RGB}{32, 64, 129}
\definecolor{darkgreen}{RGB}{0, 110, 85}
\definecolor{darkred}{RGB}{153, 0, 0}
\definecolor{graytext}{gray}{0.45}

\definecolor{evaluatorcolor}{RGB}{255,230,230}    %
\definecolor{evaluatorframe}{RGB}{180,50,50}      %
\definecolor{taskgencolor}{RGB}{230,255,230}      %
\definecolor{taskgenframe}{RGB}{50,180,50}        %
\definecolor{executioncolor}{RGB}{230,230,255}    %
\definecolor{executionframe}{RGB}{50,50,180}      %
\definecolor{lightgray}{RGB}{240,240,240}
\definecolor{darkgray}{RGB}{80,80,80}

\usepackage{pifont}
\usepackage{longtable}

\usepackage{makecell}

\usepackage{amsmath}
\usepackage{pifont}

\usepackage{booktabs}
\usepackage{multirow}
\usepackage{graphicx}
\usepackage{float}
\usepackage{caption}
\usepackage{subcaption}
\usepackage{xspace}

\usepackage[utf8]{inputenc}
\usepackage[T1]{fontenc}
\usepackage{xcolor}
\usepackage[most]{tcolorbox}
\usepackage{enumitem}
\usepackage{listings}

\definecolor{darkgray}{rgb}{0.3, 0.3, 0.3}
\definecolor{lightgray}{rgb}{0.95, 0.95, 0.95}
\definecolor{codegray}{rgb}{0.98, 0.98, 0.98}

\newtcolorbox{promptbox}[1]{
    colback=lightgray,
    colframe=darkgray,
    colbacktitle=darkgray,
    coltitle=white,
    boxrule=2pt,
    arc=0mm,
    left=10pt,
    right=10pt,
    top=10pt,
    bottom=10pt,
    fonttitle=\bfseries\large,
    title={#1},
    attach boxed title to top left={yshift=-2mm}
}

\usepackage{wrapfig}

\usepackage{algorithm}
\usepackage[noend]{algpseudocode}

\definecolor{deepred}{rgb}{0.631,0.102,0.102}
\definecolor{skyblue}{HTML}{126da2}

\definecolor{accpurple}{HTML}{A100FF}
\hypersetup{
     colorlinks = true,
     breaklinks = true,
     linkcolor = accpurple,
     urlcolor = accpurple,
     citecolor = accpurple
     }

\definecolor{orange}{rgb}{1,0.5,0}

\algnewcommand{\LineComment}[1]{\State \(\triangleright\) #1}

\title{POLARIS: Typed Planning and Governed Execution for Agentic AI in Back-Office Automation}

\author[1]{Zahra Moslemi}
  \author[2]{Keerthi Koneru}
  \author[3]{Yen-Ting Lee}
  \author[2]{Sheethal Kumar}
  \author[2]{Ramesh Radhakrishnan}

  \affil[1]{University of California, Irvine}
  \affil[2]{Center for Advanced AI, Accenture}
  \affil[3]{University of California, San Diego}

  \date{\today}
  \correspondingauthor{Keerthi Koneru (\href{mailto:keerthi.koneru@accenture.com}{keerthi.koneru@accenture.com}). All work was done during the authors’ tenure at Accenture.}

\begin{document}
\begin{abstract}
Enterprise back-office workflows require agentic systems that are auditable, policy-aligned, and operationally predictable, capabilities that generic multi-agent setups often fail to deliver. We present \textbf{\textsc{POLARIS}} (Policy-Aware LLM Agentic Reasoning for Integrated Systems), a governed orchestration framework that treats automation as typed plan synthesis and validated execution over LLM agents. A planner proposes structurally diverse, type-checked directed acyclic graphs (DAGs); a rubric-guided reasoning module selects a single compliant plan; and execution is guarded by validator-gated checks, a bounded repair loop, and compiled policy guardrails that block or route side effects before they occur. Applied to document-centric finance tasks, \textsc{POLARIS} produces decision-grade artifacts and full execution traces while reducing human intervention. Empirically, \textsc{POLARIS} achieves a micro-F1 of 0.81 on the SROIE dataset and, on a controlled synthetic suite, achieves 0.95–1.00 precision for anomaly routing with preserved audit trails. These evaluations constitute an initial benchmark for governed Agentic AI. POLARIS provides a methodological and benchmark reference for policy-aligned Agentic AI.

\end{abstract}

\maketitle
\def\Snospace~{Section }
\def\sectionautorefname{\Snospace}
\def\subsectionautorefname{\Snospace}
\def\subsubsectionautorefname{\Snospace}
\def\chapterautorefname{\Snospace}

\section{Introduction}
Enterprise back-office automation (e.g., accounts payable, contract checks) imposes requirements that generic multi-agent LLM stacks often fail to meet: actions must be auditable, operationally predictable, policy-aligned, and governed before any side effects occur, and reliably evaluable against reproducible metrics. 
In practice, untyped tool calls, best-of-$N$ prompting, and open-ended retries yield brittle pipelines with unclear provenance and no predictable service-level agreement (SLA) guarantees. We argue that meeting enterprise guarantees requires re-casting orchestration as typed, governed planning and execution where plans are type-checked and structurally diverse; selection is rubric-based and policy-aware; and execution is guarded by validators and compiled policy checks that gate side effects.

Recent progress in Agentic AI has created powerful but ungoverned multi-agent stacks. Modern LLMs now support precise language understanding and the use of controlled tools \citep{achiam2023gpt4}, enabling agentic systems for operational work. \emph{LLM-based agents} combine core reasoning with domain tools \citep{qin2024agentsurvey,wang2024agentbench}; multi-agent systems assign specialized roles across workflows \citep{gaoutil2023autogen,hong2024metagpt}; and planning-centric frameworks cast orchestration as \emph{agent-oriented planning (AOP)} with typed sub-tasks \citep{li2024aop}. In parallel, workflow-centric approaches represent execution as a directed acyclic graph (DAG) of tool calls with explicit interfaces \citep{gaoutil2023autogen,hong2024metagpt}, often following a \emph{plan-then-act} paradigm that synthesizes candidate workflows before selection \citep{gentask2025,gaoutil2023autogen}. However, enterprise deployment of such Agentic AI remains limited by the lack of typed planning, evaluation metrics, and auditable execution paths.

Applying these ideas in regulated settings is challenging: inputs are heterogeneous and legally constrained, and errors carry financial risk.
Common stacks (a) pass untyped input/output (I/O) in free-form messages, (b) treat plan generation as best-of-$N$ prompting without structural diversity guarantees, (c) lack \emph{compiled} policy checks that gate side effects, and (d) don't use past experience to guide future decisions; making governed, auditable behavior difficult to guarantee.

To overcome these inherent drawbacks, we introduce \textsc{POLARIS}, a modular orchestration framework that treats back-office automation as planning over \emph{typed} agents. A Chain-of-Action planner proposes a small set of structurally diverse type-checked DAGs; a lightweight reasoning selector chooses one plan using a rubric over compliance, safe sequencing, and parsimony. Execution is then \emph{guarded}: the validator checks gate side effects and drives a targeted parser–validator repair loop, while compiled policy guardrails (e.g., thresholds, currency rules, segregation-of-duties (SoD)) block, route, or annotate actions prior to external effects. The result is an auditable decision object and complete execution trace suitable for regulated environments (Figure~\ref{fig:architecture}). 

Relative to prior multi-agent DAG systems, POLARIS (i) enforces type-sound composition with first-class diversity constraints at plan generation (not just prompt sampling), (ii) uses a rubric-based selector that is policy-aware and emits auditable JSON decisions, and (iii) bounds latency and cost via a validator-gated repair loop that targets only failing fields before any external side-effects. In synthetic invoice suites and the SROIE benchmark, the same plan–select–act loop shows strong extraction accuracy and high-precision policy/anomaly routing while preserving end-to-end lineage.

This work contributes to the broader effort to establish reproducible benchmarks and evaluable frameworks for Agentic AI in enterprise settings. POLARIS provides a reference implementation for typed policy-aware orchestration that can be quantitatively assessed in controlled enterprise scenarios.

The remainder of this paper is organized as follows: Section~\ref{sec:relatedwork} reviews prior advances in agentic process automation, benchmarking, and programmatic prompting. Section~\ref{sec:prelims} introduces the enterprise task setting, notation, and agent taxonomy used throughout this work. Section~\ref{sec:method} formalizes the POLARIS framework, covering typed plan synthesis, rubric-based reasoning, and dependency-aware execution. Sections~\ref{sec:repair}--\ref{sec:policy} describe the validator-gated repair loop, policy guardrails, and risk control mechanisms. Section~\ref{sec:experiments} presents empirical evaluations and benchmark comparisons, while Section~\ref{sec:conclusion} concludes with implications for governed, auditable agentic automation and future benchmark development.
\section{Related Work}
\label{sec:relatedwork}

LLM-based back-office automation spans general orchestration frameworks and domain-specific systems. \textbf{ProAgent} reframes Robotic Process Automation as Agentic Process Automation by having an LLM synthesize executable workflows and coordinate sub-agents for data-dependent branches~\citep{ye2023proagent}. \textbf{SmartFlow} complements this with vision and LLM planning to operate changing enterprise UIs without fragile selectors~\citep{jain2024smartflow}. Domain-aligned approaches harden governance: \textbf{FinRobot} introduces finance-focused Generative Business Process AI Agents that map intents to ERP tasks via rule-constrained planning, Chain-of-Actions execution, and human-in-the-loop for auditability~\citep{yang2025finrobot}; \textbf{Agent-S} formalizes the execution of the SOP with a state-decision planner, an action executor for API/user steps, and shared execution memory for fault recovery~\citep{kulkarni2025agents}; and an \textbf{RL-guided multi-agent parser} learns document classification, schema induction, and iterative extraction to handle layout drift in invoices and POs~\citep{amjad2025agentic}. Although not enterprise-specific, \textbf{Generative Agents} contribute reusable patterns for memory, retrieval, and reflection that stabilize multi-step behaviors~\citep{park2023generative}. 

\textbf{Benchmarks and evaluation.} Recent evaluation efforts emphasize \emph{process-aware} benchmarking beyond final task success. \textbf{AgentBench} and \textbf{AgentBoard} introduce structured multi-environment evaluations with fine-grained analytical metrics for LLM agents~\citep{liu2023agentbench,ma2024agentboard}, while \textbf{BrowserGym} and \textbf{WebArena} provide reproducible web-based environments to test planning, tool use, and long-horizon reasoning~\citep{browsergym2024,zhou2023webarena}. Governance-focused frameworks such as \textbf{Safe Agents}~\citep{huang2024safeagents} and rubric-based evaluation methods such as \textbf{LLM-as-Judge}~\citep{zheng2024llmasjudge} underscore the importance of accountability and transparent decision traces—key motivations for POLARIS's policy-aware orchestration. Our work focuses on typed planning, policy-aware selection, and validator-gated repair producing auditable traces, areas not emphasized in UI/web benchmarks.

\textbf{Programmatic prompting and planning priors.} Beyond conventional prompt templates, \textbf{DSPy} compiles declarative LLM pipelines into self-improving modules and provides structured priors that enable optimizable, type-constrained plan synthesis~\citep{khattab2023dspy}. \textbf{AutoGen} demonstrates scalable multi-agent conversations with coordination patterns, but lacks type contracts or compiled policy guardrails~\citep{wu2023autogen}. In broader multi-agent planning, \textbf{Voyager}~\citep{wang2023voyager} shows how exemplar-guided exploration and continual learning improve long-horizon agent behavior. \textsc{POLARIS} unifies these strengths by enforcing \emph{typed planning}, \emph{policy compliance}, and \emph{bounded repair loops}, establishing a governance layer critical for enterprise-grade auditability and reliability.

\section{Preliminaries}
\label{sec:prelims}

We target document-centric enterprise workflows (e.g., invoice processing) where heterogeneous inputs—PDFs, emails, prompts, or API events—must yield \emph{decision-grade} artifacts such as validated records, approvals, or audit reports. 
Let $\mathcal{X}$ denote raw inputs and $\tau\!\in\!\mathcal{T}$ be the normalized task record summarizing the metadata and required fields.

\paragraph{Typed agents and I/O contracts.}
POLARIS maintains a library $\mathcal{A}=\{A_1,\dots,A_n\}$ of GPT-backed components, each specifying a capability description, typed I/O schemas, pre/postconditions, and policy or risk tags (eligibility, side effects, segregation-of-duties). 
Agents are grouped as: \textbf{Normalizer} (\texttt{InputNormalizer}); \textbf{Planning \& Selection} (\texttt{CoAPlanner}, \texttt{ReasoningAgent}); \textbf{Data Extractors} (\texttt{DocumentParser}, \texttt{DataValidator}); \textbf{Data Processors} (\texttt{RecordMatcher}, \texttt{PolicyRetrieval}, \texttt{AnomalyDetection}, \texttt{RiskControl}, \texttt{APIAccess}, \texttt{Scheduler}); and \textbf{Reconciliation} (\texttt{Approval}, \texttt{ReportGenerator}). 

\paragraph{Plans as typed DAGs.}
A plan $p$ is a directed acyclic graph $G_p=(V_p,E_p)$ of agent invocations whose edges are type-checked against declared I/O contracts. 
Feasibility requires: (i) input compatibility, (ii) compliance ordering (e.g., \textit{parse}$\!\rightarrow$\textit{validate}$\!\rightarrow$\textit{risk/approval}), and (iii) adherence to concurrency limits. 
Execution follows \emph{completion-based} semantics—nodes become eligible when all parents finish—allowing safe fan-out and aggregation of partial evidence. 

\paragraph{Compliance and anomaly primitives.}
Policy checks detect unknown vendors, threshold breaches, and currency mismatches (Section~\ref{sec:policy}). 
Anomalies use a robust vendor-wise score:
\[
z_{\mathrm{MAD}}(x;v)=\frac{|x-\mathrm{median}_v|}{1.4826\cdot\mathrm{MAD}_v},
\]
flagged if $z_{\mathrm{MAD}}>k_{\mathrm{mad}}$ (default $3.5$), with cohort/global fallback and date-sanity validation (Section~\ref{sec:anomaly}). \textit{Additional details, agent role definitions, Table~\ref{tab:ops-agents-full} appear in Appendix~\ref{app:prelims_full}.}

\section{Method}
\label{sec:method}
\begin{figure*}[t]
  \centering
  \includegraphics[width=0.7\textwidth, keepaspectratio]{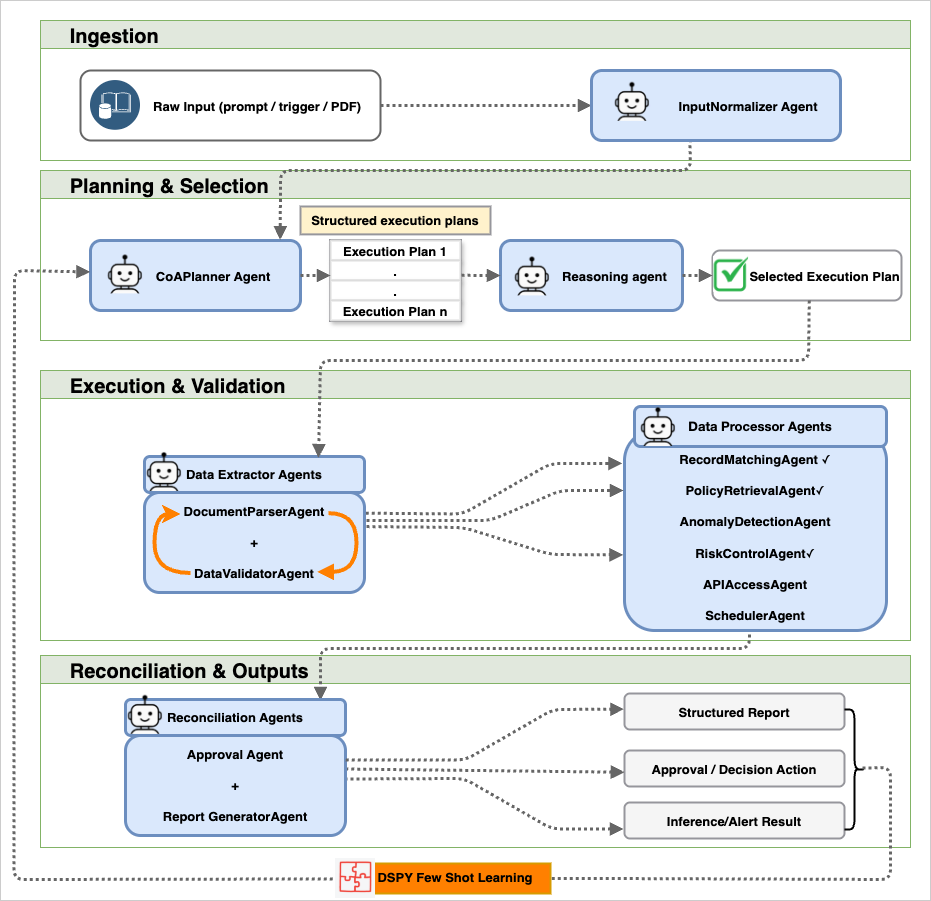}
  \caption{\textbf{Overview of POLARIS orchestration with governed plan selection.} Raw inputs (prompts/triggers/PDFs) are normalized by the \texttt{InputNormalizerAgent}. The \texttt{CoAPlannerAgent} proposes multiple structured execution plans; a \texttt{Reasoning Agent} scores them and selects a single plan. The chosen plan executes as a DAG: a \texttt{DocumentParserAgent} and \texttt{DataValidatorAgent} run with a bounded repair loop (orange) to improve extractions, after which only the required \textit{Data Processor Agents} are invoked (e.g., \texttt{RecordMatching}, \texttt{PolicyRetrieval}, \texttt{AnomalyDetection}, \texttt{RiskControl}, \texttt{APIAccess}). \textit{Reconciliation Agents} (\texttt{Approval}, \texttt{ReportGenerator}) produce decision-grade outputs (Structured Report, Approval/Decision Action, Inference/Alert Result). Dashed lines indicate data/control flow across stages. DSPy few-shot learning supplies planning priors.}
  \label{fig:architecture}
\end{figure*}
\subsection{POLARIS Overview}
POLARIS operationalizes the above as a \emph{plan–select–act} loop over typed agents (Figure~\ref{fig:architecture}). It modularizes normalization, plan synthesis, reasoning-driven selection, guarded execution, and continuous feedback—each stage auditable and policy-aligned.

\subsection{(1) Normalization}
\textit{This stage converts heterogeneous input formats from the enterprise into a canonical JSON record, enabling type verification and policy tagging for reproducible downstream execution.}

\textbf{InputNormalizerAgent} maps raw inputs $x\!\in\!\mathcal{X}$ (files, free text, events) to a canonical task record $\tau=f_{\mathrm{norm}}(x)$ consumed by downstream agents.  
It handles diverse inputs (\textit{schedules, prompts, PDFs, emails, API events}), categorizes, and normalizes them for consistent downstream processing.

Concretely, the agent is an \texttt{Autogen ConversableAgent} prompted with a strict JSON schema and returns a single JSON object with fields:
\begin{itemize}
  \item \texttt{task\_type}: coarse task category (e.g., \textit{document\_parsing}, \textit{user\_command}, \textit{event\_triggered});
  \item \texttt{input\_format}: acquisition mode (e.g., \textit{file}, \textit{text}, \textit{file+instruction}, \textit{event});
  \item \texttt{file\_name}: original filename if present, otherwise empty;
  \item \texttt{file\_type}: MIME/extension (e.g., \textit{pdf}, \textit{png}, \textit{eml}, \textit{csv}) or \textit{none};
  \item \texttt{timestamp}: ISO-8601 time associated with $x$ (ingest or document time);
  \item \texttt{origin}: source identifier (e.g., \textit{email}, \textit{upload}, \textit{chat});
  \item \texttt{instruction}: user/system instruction if supplied;
  \item \texttt{meta}: auxiliary key–value pairs (e.g., locale, BU, region, channel).
\end{itemize}

The implementation enforces  
(i) schema-guided prompting,  
(ii) model invocation via \texttt{generate\_reply}, and  
(iii) deterministic JSON extraction—the last JSON object in the response is regex-parsed.  
In case of failure to recover a valid JSON, the agent returns a structured error payload with the raw model response, enabling safe fallback or escalation without side effects.  
This normalization removes input heterogeneity before planning, ensuring reproducible type checks and policy attachment.

\subsection{(2) Candidate Plan Synthesis (CoAPlanner)}
\label{sec:coaplanner}

\textbf{Goal.}
Given a normalized task record $\tau$, the planner must propose a \emph{set} of $K$ \emph{typed}, \emph{feasible}, and \emph{pairwise distinct} workflow hypotheses
\[
\Pi(\tau)=\{p_1,\dots,p_K\},\qquad p_k\in\mathcal{P},
\]
where each $p_k$ is a directed acyclic graph (DAG) of agent invocations whose edges respect the declared input/output (I/O) types and policy tags (Section~\ref{sec:prelims}). 
Plan synthesis is modeled as diversification under type constraints guided by model priors.

\paragraph{Core reasoning steps.}
Internally, the CoAPlanner performs three sequential reasoning steps before generating candidate DAGs:
\begin{itemize}
  \item \textbf{Interpret Task Intent}: Derive the required outcome from the parsed \texttt{task\_type} of $\tau$;
  \item \textbf{Select Required Agents}: identify task-specific agents relevant to execution;
  \item \textbf{Sequence Agent Calls}: order agents by dependencies and conditional triggers.
\end{itemize}

\paragraph{Typed, constraint-guided generation.}
We view the planner as a conditional generator $G_\theta(p\mid\tau,\mathcal{E})$ on typed DAGs, where $\mathcal{E}$ is a small exemplar bank.  
Each draw $p\!\sim\!G_\theta$ must satisfy  
\emph{feasibility} (type-compatible edges; intrinsic orderings such as \textit{parse}$\!\rightarrow$\textit{validate}$\!\rightarrow$\textit{risk/approval})  
and \emph{minimality} (no superfluous stages).  
Feasibility is enforced at generation by emitting node signatures and checking edge composition against I/O contracts; minimality is encouraged through a brevity prior in the planner’s scoring.

\paragraph{Diversity as a first-class constraint.}
Rather than sample duplicates and prune, we **enforce diversity during generation.**  
Let $\sim$ denote an equivalence relation identifying DAGs with the same ordered “core chain” and dependency structure.  
The planner returns $\Pi(\tau)$ such that $p_i\not\sim p_j$ for all $i\!\ne\!j$.  
Practically, candidates must differ in agent set, edge structure, or stage ordering while maintaining type soundness.  
Formally, the planner solves:
\begin{equation}
\max_{\Pi(\tau)} \sum_{p\in\Pi(\tau)} s_\theta(p;\tau)
\quad\text{s.t.}\quad
p\!\in\!\mathcal{F}(\tau)\ \forall p,\; p_i\!\not\sim\!p_j\ \forall i\neq j.
\end{equation}

where $s_\theta$ is the planner’s internal plausibility score and $\mathcal{F}(\tau)$ the feasible set.

\paragraph{Domain priors via few-shot planning.}
To bias generation toward policy-consistent plans, we instantiate $\mathcal{E}$ (few-shot memory) using \textbf{DSPy (Declarative Self-improving Python)}, which compiles LLM prompts into structured, trainable programs.  
DSPy encodes reusable planning modules that are optimizeable and updateable.
For CoAPlanner, $\mathcal{E}$ includes compact labeled scenarios (positive/negative, anomaly exemplars, approval thresholds) describing the ideal agent chains and rationales.  
These exemplars are compiled by DSPy into a reusable planning program conditioning $G_\theta$ during generation.  
When producing its $K$ DAGs, CoAPlanner biases sampling toward successful chain patterns (e.g., \textit{parse}→\textit{validate}→\textit{risk/approve}) while exploring alternative edge structures. This yields efficient and robust few-shot planning.

The Reasoner later evaluates these DSPy-guided candidates, ensuring that only type- and policy-sound DAGs advance.  
Thus DSPy supplies CoAPlanner with a structured prior narrowing the search space to high-quality, diverse, interpretable $K$-plan proposals.

\paragraph{Policy-aware invariants.}
Certain invariants are hard-coded: invoices with vendors must retrieve policy context before irreversible actions; “month-end” workflows must stage scheduling before report emission; and approval sinks must trail validation and risk control. These invariants restrict $\mathcal{F}(\tau)$ to type- and policy-sound plans.

\subsection{(3) Reasoning-Based Selection (ReasoningAgent)}
\label{sec:selection}

\textbf{Objective.}
Given a diversified set $\Pi(\tau)=\{p_1,\dots,p_K\}$ of typed plans (Section~\ref{sec:coaplanner}),  
the selector chooses a single plan
\begin{equation}
   \hat p = \arg\max_{p\in\Pi(\tau)} U(p;\tau)
\end{equation}
with
\begin{align}
U(p;\tau) &= w_1\,\text{compliance}(p) + w_2\,\text{sequencing}(p) \\
&\quad + w_3\,\text{parsimony}(p) + w_4\,\text{prior}(p).
\end{align}

The four rubric terms are:
(i) \textbf{compliance}: verifies admissible agents and contract/policy eligibility;  
(ii) \textbf{sequencing}: evaluates ordering quality, favoring safe sequences (\textit{parse}$\!\rightarrow$\textit{validate}$\!\rightarrow$\textit{risk/approval}) while penalizing unsafe permutations;  
(iii) \textbf{parsimony}: prefers minimal non-redundant step sets; and  
(iv) \textbf{prior}: planner-assigned self-consistency score.  
Weights $(w_1,w_2,w_3)$ are fixed in this work.

\paragraph{Selector as a constrained reasoner.}
We instantiate the selector as a lightweight \emph{reasoning} LLM that (a) evaluates rubric terms against the typed plan specification and (b) emits a \emph{structured} decision: a plan index and a short justification. Using the normalized record, the candidates $K$, and the rubric, it returns a JSON object containing \texttt{chosen\_index} and \texttt{reason}.

Plans that fail any hard constraint (policy or type) are discarded before scoring.

Reasoning-tuned LLMs improve \textit{constraint awareness}, \textit{calibrated comparison}, and \textit{structured output control}. They maintain policy/order invariants, apply rubric ties rationally, and emit strict JSON enabling fail-fast rejection and auditable decisions. Importantly, the selector evaluates only among pre-validated candidates without altering types or adding actions.

\begin{figure}[t]
  \centering
  \includegraphics[width=0.6\columnwidth]{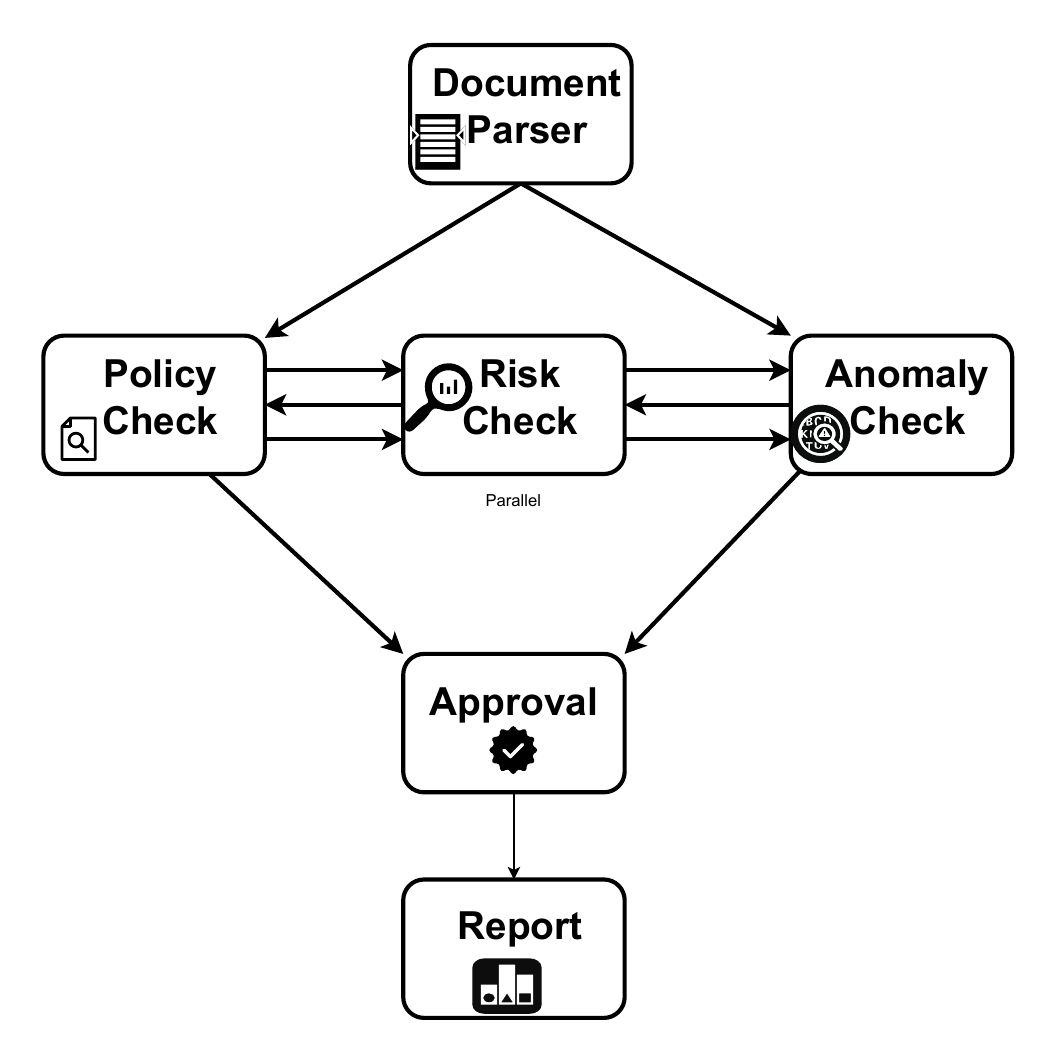}
  \caption{\textbf{Dependency-aware parallel scheduler.}
  The scheduler converts the planner’s ordered list of agents into a DAG that enables safe parallelism.
  \texttt{DocumentParser} runs first; middle-stage checks run concurrently; sinks wait for all upstream checks.}
  \label{fig:dependency}
\end{figure}

\subsection{(4) Guarded Execution with a Dependency-Aware Scheduler}
Given the selected plan $\hat p$, the execution proceeds as a dependency graph with completion-based semantics. The scheduler converts CoAPlanner’s ordered stages into a typed DAG and runs ready nodes in parallel while honoring dependencies (Figure~\ref{fig:dependency}). 

\paragraph{Execution model.}
We maintain a readiness set $R$ of nodes whose predecessors have been successfully completed. At each step, the scheduler dispatches all nodes in $R$ subject to resource limits; completions update downstream dependencies and refresh $R$. 

The resulting parallel execution pattern is shown in Figure~\ref{fig:dependency}.

\paragraph{Dependency rules (enforced invariants).}
\begin{itemize}
  \item \textbf{Parser first:} any agent that consumes extracted fields must wait for \texttt{DocumentParser}.
  \item \textbf{Middle parallelization:} after parsing, independent checks (\texttt{PolicyRetrieval}, \texttt{RiskControl}, \texttt{AnomalyDetector}, \texttt{RecordMatcher}) run concurrently.
  \item \textbf{Aggregate sinks last:} \texttt{Approval} and \texttt{ReportGenerator} execute only after all upstream checks succeed.
\end{itemize}

\paragraph{Dynamic prerequisites.}
Edges may be injected at runtime based on intermediate signals. For instance, if \texttt{AnomalyDetector} emits a high score, the scheduler adds a prerequisite edge to an \texttt{ExtraVerification} node; conversely, low risk may bypass optional reconciliation. These dynamic guards preserve safety while minimizing latency.

\subsection{(5) Validator-Gated Repair Loop}
\label{sec:repair}

Document parsing is the dominant failure mode in back-office flows: OCR/layout models can miss fields, mis-read dates/currencies, or drift under noisy scans. Executing brittle parses propagates policy errors. We therefore insert a bounded, validator-gated repair loop to improve extractions before irreversible actions.

\paragraph{Trigger.}
Let $\hat z$ be the parser output (fields with confidences), and $V(\hat z)$ denote validator checks (schema completeness, cross-field arithmetic, date logic, external consistency). The loop is entered if:
(i) required field missing; or (ii) field confidence $<\tau_c$; or (iii) rule violation $V(\hat z)=\text{fail}$.

Instead of re-parsing wholesale, we issue localized repair actions guided by validator feedback: ROI hints (zones for totals/dates/vendors), OCR/layout cues, schema prompts, and string normalization.
 
Repairs are additive; only implicated fields are re-requested.

\paragraph{Termination and guarantees.}
The loop runs for at most $L_{\max}$ iterations (typically $2\!\sim\!3$) and terminates when $V(\hat z)=\text{pass}$ or the budget is exhausted. If it exits by budget, control escalates for human review or a safe lower-coverage path (no approvals, full audit trace).

\paragraph{Scheduler integration.}
The repair loop lies in the critical path before mid-pipeline checks. After completion, repaired fields seed the execution store, skipping redundant \texttt{DocumentParser}/\texttt{DataValidator} stages, and preserving parallelism downstream.

\begin{algorithm}[t]
\caption{\textbf{Validator-Gated Repair Loop (bounded)}}
\label{alg:repair}
\begin{algorithmic}[1]
\Require normalized record $\tau$, parser $P$, validator $V$, threshold $\tau_c$, budget $L_{\max}$
\State $\hat{z} \gets P(\tau)$ \Comment{initial parse}
\For{$\ell = 1$ \textbf{to} $L_{\max}$}
  \State $r \gets V(\hat{z})$ \Comment{schema / cross-field checks; $r{=}\text{pass}$ and $\minconf(\hat{z}) \ge \tau_c$}
  \If{$r = \text{pass}$}
    \State \textbf{break} \Comment{accept repaired extraction}
  \EndIf
  \State $\mathcal{A} \gets \textsc{Explain}(r, \hat{z})$ \Comment{which fields/rules failed?}
  \State $\hat{z} \gets \textsc{Merge}(\hat{z},\, P(\tau;\mathcal{A}))$ \Comment{re-parse + overwrite}
\EndFor
\If{$V(\hat{z}) = \text{pass}$}
  \State \Return $\hat{z}$ \Comment{seed downstream agents}
\Else
  \State \Return $\textsc{Fallback}(\hat{z})$ \Comment{safe path}
\EndIf
\end{algorithmic}
\end{algorithm}

\subsection{(6) Policy Guardrails and Risk Control}
\label{sec:policy}
\textbf{PolicyRetrieval} queries a lightweight policy store keyed by vendor, sector and currency, returning (i) existence flag, (ii) matched vendor identity, and (iii) the vendor’s policy record (threshold, currency, terms).  
\textbf{RiskControl} compiles these clauses into executable checks and gates unsafe actions: hard violations halt; soft ones escalate with machine-readable justifications.

\paragraph{Violation definition.}
Let $E(v)$ denote the existence of the vendor $v$, 
$a(i)$ the total invoice for the invoice $i$, $\tau(v)$ the threshold for the vendor $v$, $c(i)$ the inferred currency
, and $c^*(v)$ the policy currency
A violation occurs if any holds:
\begin{enumerate}
  \item \textbf{Unknown vendor:} $\neg E(v)$.
  \item \textbf{Threshold breach (unapproved):} $E(v)\land a(i)>\tau(v)$ and no attached approval artifact.
  \item \textbf{Currency mismatch:} $E(v)\land c(i)\neq c^*(v)$.
\end{enumerate}

\paragraph{Evaluation.}
We summarize policy decisions by $(\mathrm{TPV},\mathrm{FPV},\mathrm{FNV},\mathrm{TNV})$:
\begin{itemize}
  \item \textbf{TPV} (True Positive Violation): the system correctly flags a true violation;
  \item \textbf{FPV} (False Positive Violation): the system flags violation when none exists;
  \item \textbf{FNV} (False Negative Violation): the system misses a true violation;
  \item \textbf{TNV} (True Negative Violation): the system correctly identifies compliance.
\end{itemize}
We report precision/recall/F1 scores for policy compliance.

\paragraph{Extending the policy layer.}
Beyond core checks, the layer supports: (i) duplicate detection (same $(\mathrm{vendor},\mathrm{invoice\_number})$ within a lookback window $T$); (ii) black/whitelists; (iii) graded risk scoring, 
$s(i)=\sum_k w_k\,\mathbf{1}\{\mathrm{rule}_k(i)\}$ for tiered routing (auto-approve / review / block); and 
(iv) enforcement of provenance that requires PO, receiving, and approval artifacts.
Missing provenance automatically registers as a violation.

\subsection{(7) Anomaly Detection and Handling}
\label{sec:anomaly}
\textbf{Signal construction.}
For each vendor $v$, we maintain baselines over the amount fields and compute 
$z_{\mathrm{MAD}}(x;v)=|x-\mathrm{median}_v|/(1.4826\cdot \mathrm{MAD}_v)$.
An item is anomalous if $z_{\mathrm{MAD}}>k_{\mathrm{mad}}$ (default $3.5$). 
When vendor data are sparse ($\mathrm{MAD}_v{=}0$ or few points), we substitute cohort (sector/currency) or global statistics to avoid false spikes.
Additionally, we apply \emph{date sanity} rules: any issue/due/payment date strictly after ``today'' is anomalous.

\begin{table*}[t]
\centering
\caption{\textbf{Extraction} by scenario (required fields: invoice \#, date, vendor, total).}
\label{tab:extract}
\resizebox{0.85\linewidth}{!}{
\begin{tabular}{lrrrrrrrr}
\hline \rule{0pt}{1.1\normalbaselineskip}
\textbf{Scenario} & \textbf{total} & \textbf{TP} & \textbf{FP} & \textbf{FN} & \textbf{TN} & \textbf{precision} & \textbf{recall} & \textbf{f1} \\
\hline \rule{0pt}{1.1\normalbaselineskip}
\textit{Violation–Unknown Vendor (VU)} & 40 & 35 & 1 & 0 & 4 & 0.9722 & 1.0000 & 0.9859 \\
\textit{Violation–Layout Drift and Noise (VL)}            & 40 & 23 & 5 & 0 & 12 & 0.8214 & 1.0000 & 0.9020 \\
\textit{Compliant--Clean Set (CC)}            & 40 & 29 & 1 & 2 & 8 & 0.9667 & 0.9355 & 0.9508 \\
\textit{Compliant–Month-End Batch (CM)}         & 40 & 34 & 0 & 2 & 4 & 1.0000 & 0.9444 & 0.9714 \\ 
\hline \rule{0pt}{1.1\normalbaselineskip}
\textbf{TOTAL} & 160 & 121 & 7 & 4 & 28 & \textbf{0.9453} & \textbf{0.9680} & \textbf{0.9565} \\
\hline
\end{tabular}}
\end{table*}

\paragraph{Routing playbook.}
A downstream playbook executes \emph{enrich $\rightarrow$ classify severity $\rightarrow$ route $\rightarrow$ act $\rightarrow$ close}:
(i) enrich with policy context and prior incidents; 
(ii) classify the severity;
(iii) route to \texttt{APIAccess} holds, ticketing, or human review; 
(iv) act (notify vendor, request artifacts, schedule re-checks); 
(v) close with a structured disposition. 
Outcomes refresh vendor/cohort baselines and dashboards, preventing drift.  
This design keeps anomaly detection statistically minimal (see Section~\ref{sec:prelims}) but operationally governed and auditable.

\subsection{(8) Outputs, Auditability, and Feedback}
\label{sec:outputs}

\paragraph{Decision-grade artifacts.}
For each input, POLARIS emits: (i) a validated JSON record with fields, confidences, and type signatures; (ii) a decision object (approve/hold/reject) with structured rationales tied to policy clauses and anomaly evidence; (iii) alerts and escalation targets; and (iv) a complete execution trace (selected DAG, per-node I/O, repair actions, and timing). These artifacts enable replay, independent verification, and seamless system ingestion.

\paragraph{Auditability by construction.}

POLARIS logs traces and summaries—what plan ran, why it was chosen, what failed, and what was repaired. These logs update (i) planning priors (few-shot exemplars) and (ii) anomaly/policy baselines. Improvement is governed: exemplars, not opaque prompts, are updated, preserving predictability and reviewability.

\begin{figure}[t]
  \centering
  \includegraphics[width=0.85\linewidth]{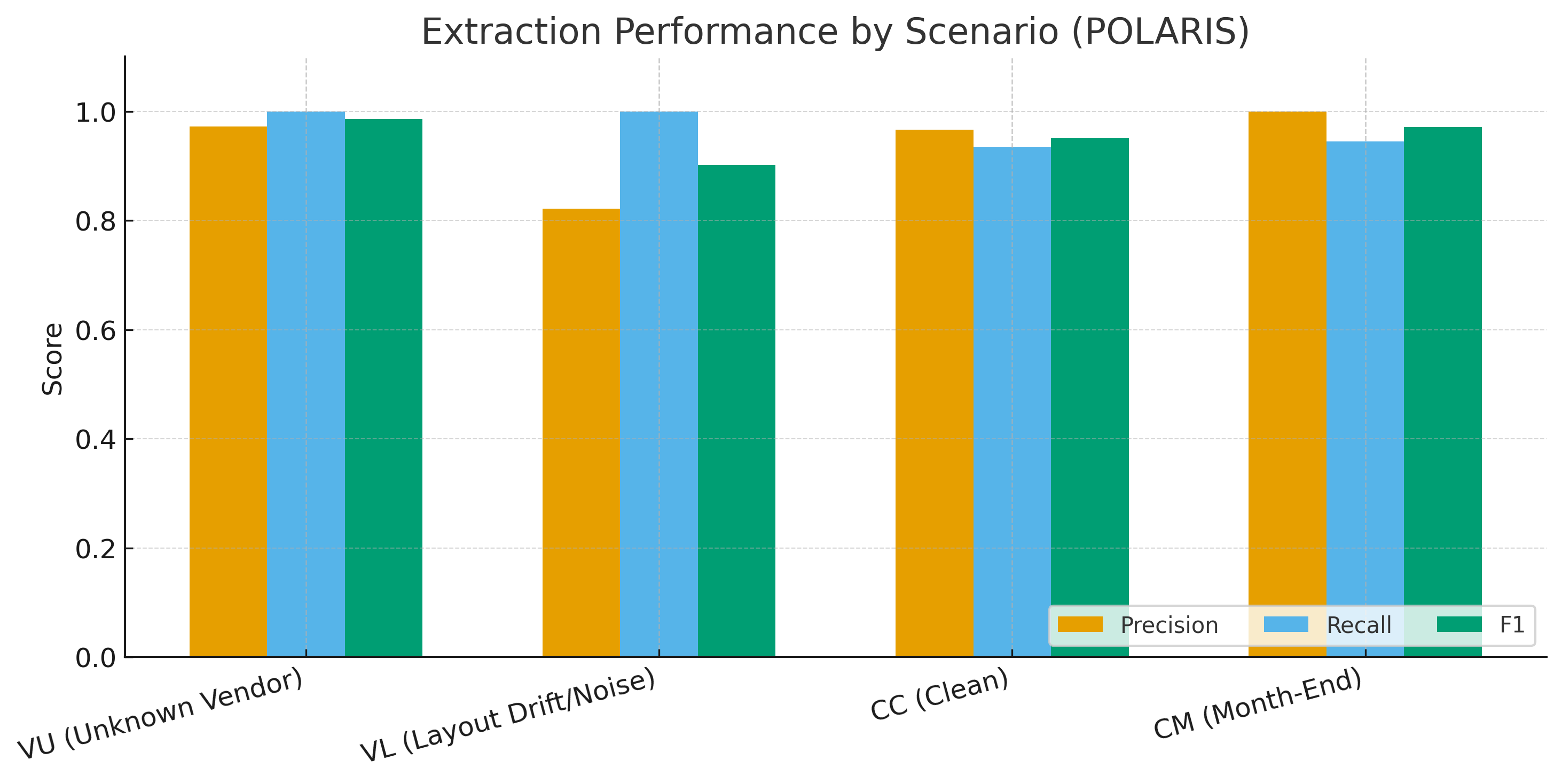}
  \caption{Extraction performance by scenario for \textbf{POLARIS}. Grouped bars show Precision, Recall, and F1 across VU (Unknown Vendor), VL (Layout Drift/Noise), CC (Clean), and CM (Month-End). The plot highlights high recall under noise (VL) with a precision dip, and strong balance on clean/compliant settings.}
  \label{fig:extraction-scenarios}
\end{figure}

\section{Evaluation and Benchmark Results}
\label{sec:experiments}

We evaluate POLARIS as a benchmarking framework for governed Agentic AI in enterprise workflows, using (i) a controlled synthetic suite that stresses policy governance, anomaly handling, and scheduler behavior (in Table~\ref{tab:synthetic-design}), and (ii) the SROIE benchmark for document-centric extraction.

\paragraph{Models.}
Unless otherwise stated, CoAPlanner is instantiated with \textbf{GPT-4o} (diversity-oriented plan synthesis under type constraints), while ReasoningAgent is a \textbf{GPT-5} reasoning model configured for strict JSON output and fail-fast selection. This mirrors our design: open-ended planning with a high-throughput multimodal model, followed by compact, constraint-aware comparison with a reasoning-tuned model.

\begin{figure*}[t]
  \centering
  \includegraphics[width=\textwidth]{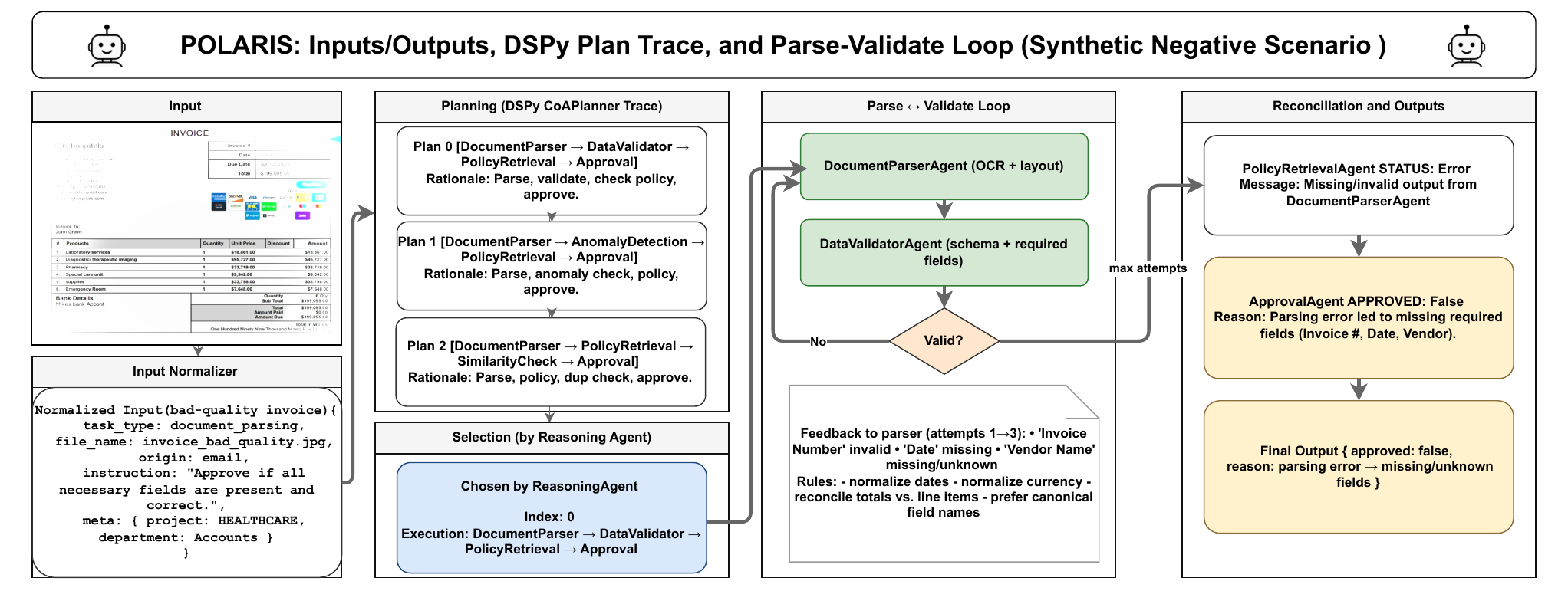}
  \caption{\textbf{POLARIS: Inputs/Outputs, DSPy plan trace, and validator-gated repair loop (synthetic negative scenario VL).}
  Left: a low-quality invoice is normalized by the \texttt{InputNormalizer} into a canonical JSON record.
  Middle–left: the \texttt{CoAPlanner} proposes multiple typed plans; the \texttt{ReasoningAgent} selects one (shown) that executes \texttt{DocumentParser} $\rightarrow$ \texttt{DataValidator} $\rightarrow$ \texttt{PolicyRetrieval} $\rightarrow$ \texttt{Approval}.
  Middle–right: a bounded parser$\leftrightarrow$validator loop issues targeted feedback (e.g., fix missing/invalid invoice number, date, vendor; normalize currency/dates) until validation passes or the attempt budget is exhausted.
  Right: reconciliation shows a policy retrieval error due to missing/invalid parsed fields; the \texttt{ApprovalAgent} returns \texttt{approved: false} with a structured rationale and final output object.}
  \label{fig:polaris-parse-validate-negative}
\end{figure*}

\begin{table}[t]
\centering
\small
\caption{\textbf{Synthetic suite and labeling rules.} Each scenario contains 10 invoices (40 total). Policy violation ground truth is \emph{vendor not found in policy DB}. Anomaly ground truth uses a robust z-score with MAD: $|x-\mathrm{median}(X)|/(1.4826\cdot\mathrm{MAD}(X))>k_{\text{mad}}$; we use $k_{\text{mad}}{=}3.5$.}
\label{tab:synthetic-design}
\begin{tabular}{p{5.6cm}p{8.8cm}}
\hline
\textbf{Scenario} & \textbf{Description / Purpose} \\
\hline
\textit{Compliant--Clean Set (CC)} & Clean invoices; vendors present in policy DB; no anomalies expected. \\
\textit{Compliant--Month-End Batch (CM)} & Day-30 month-end batch; parsing/policy clean; scheduler must trigger. \\
\textit{Violation--Unknown Vendor (VU)} & Vendors intentionally absent from policy DB $\Rightarrow$ policy violation expected. \\
\textit{Violation--Layout Drift \& Noise (VL)} & Noisy/misaligned scans $\Rightarrow$ stresses extraction \& anomaly robustness. \\
\hline
\textbf{SROIE} & Standard 4-field extraction (Company, Address, Date, Total). \\
\hline
\end{tabular}
\end{table}

\subsection{Datasets and Simulation Design}

\paragraph{Setup.}
All runs use the plan–select–act loop with $K{=}5$ diversified plans, the constrained \emph{ReasoningAgent} selector, and the \emph{validator-gated repair loop} with budget $L_{\max}{=}3$. Month-end jobs enforce \texttt{Scheduler}$\rightarrow$\texttt{ReportGenerator}. Anomaly detection uses vendor-wise MAD baselines with $k_{\text{mad}}{=}3.5$ plus a strict future-date rule.

\begin{table}[t]
\centering
\caption{\textbf{Policy} (violation detection). Metrics are undefined (\textemdash{}) where no positive cases exist.}
\label{tab:policy}
\resizebox{0.75\linewidth}{!}{
\begin{tabular}{lrrrrrrr}
\hline \rule{0pt}{1.1\normalbaselineskip}
\textbf{Scenario} & \textbf{TPV} & \textbf{FPV} & \textbf{FNV} & \textbf{TNV} & \textbf{precision} & \textbf{recall} & \textbf{f1} \\
\hline \rule{0pt}{1.1\normalbaselineskip}
\textit{VU} & 8 & 1 & 0 & 0 & 0.8889 & 1.0000 & 0.9412 \\ \rule{0pt}{1.1\normalbaselineskip}
\textit{VL} & 1 & 1 & 2 & 2 & 0.5000 & 0.3333 & 0.4000 \\ \rule{0pt}{1.1\normalbaselineskip}
\textit{CC} & 0 & 0 & 0 & 5 & \textemdash{} & \textemdash{} & \textemdash{} \\ \rule{0pt}{1.1\normalbaselineskip}
\textit{CM} & 0 & 0 & 0 & 8 & \textemdash{} & \textemdash{} & \textemdash{} \\ 
\hline \rule{0pt}{1.1\normalbaselineskip}
\textbf{TOTAL} & 9 & 2 & 2 & 15 & \textbf{0.8182} & \textbf{0.8182} & \textbf{0.8182} \\
\hline \rule{0pt}{1.1\normalbaselineskip}
\end{tabular}}
\end{table}

\subsection{Synthetic Results (40 Invoices)}

\paragraph{Field extraction.}
Across scenarios, \textsc{POLARIS} achieves high recall with strong overall balance (Table~\ref{tab:extract} and Figure~\ref{fig:extraction-scenarios}): precision/recall/F1 of 0.945/0.968/0.957. In VU (unknown vendors), extraction is accurate despite policy violations (precision 0.972, recall 1.000), which shows the method’s ability to separate parsing quality from downstream governance. The hardest case, VL (noisy/misaligned scans), preserves recall 1.000 with a precision dip (0.821), indicating that robustness favors coverage under noise. Clean conditions perform as expected: CC yields high balance (precision 0.967, recall 0.936) and CM achieves perfect precision with slightly reduced recall (precision 1.000, recall 0.944). An end-to-end trace of a noisy run appears in Figure~\ref{fig:polaris-parse-validate-negative}.

\paragraph{Policy governance.}
Unknown-vendor violations (VU) are surfaced with high precision (Table~\ref{tab:policy}); the residual recall loss in \textit{VL} is attributable to upstream noise.

For positive scenarios (CC, CM), no policy violations are present by design; thus, there are no positive cases to score and precision/recall/F1 are undefined. When violations exist (VU, VL), precision/recall/F1 are computed normally.

\paragraph{Anomaly detection.}
Vendor-wise MAD baselines yield perfect precision and high recall overall (Table~\ref{tab:anom}).

CC has no injected anomalies, so precision/recall/F1 are N/A. VL contains layout noise, causing one FN due to sparse vendor baselines; our MAD detector return to cohort/global stats, yielding high precision with a small recall drop. VU shows no amount outliers by construction (policy, not statistical, violations), hence 0 for anomaly metrics while date checks remain active.

\begin{table}[t]
\centering
\caption{\textbf{Anomaly} (MAD, $k_{\text{mad}}{=}3.5$) and date checks. Metrics are undefined (\textemdash{}) where no positive cases exist.}
\label{tab:anom}
\resizebox{0.75\linewidth}{!}{
\begin{tabular}{lrrrrrrrr}
\hline \rule{0pt}{1.1\normalbaselineskip}
\textbf{Scenario} & \textbf{total} & \textbf{TP} & \textbf{FP} & \textbf{FN} & \textbf{TN} & \textbf{precision} & \textbf{recall} & \textbf{f1} \\
\hline \rule{0pt}{1.1\normalbaselineskip}
\textit{VU} & 8 & 0 & 0 & 0 & 8 & \textemdash{} & \textemdash{} & \textemdash{} \\ \rule{0pt}{1.1\normalbaselineskip}
\textit{VL} & 6 & 5 & 0 & 1 & 0 & 1.0000 & 0.8333 & 0.9091 \\ \rule{0pt}{1.1\normalbaselineskip}
\textit{CC} & 3 & 0 & 0 & 0 & 3 & \textemdash{} & \textemdash{} & \textemdash{} \\ \rule{0pt}{1.1\normalbaselineskip}
\textit{CM} & 9 & 4 & 0 & 0 & 5 & 1.0000 & 1.0000 & 1.0000 \\
\hline \rule{0pt}{1.1\normalbaselineskip}
\textbf{TOTAL} & 26 & 9 & 0 & 1 & 16 & \textbf{1.0000} & \textbf{0.9000} & \textbf{0.9474} \\
\hline
\end{tabular}}
\end{table}

\subsection{SROIE (4-Field Extraction)}
We apply the same parser–validator pattern (contrast upscaling $\rightarrow$ OCR $\rightarrow$ LLM with validation feedback). POLARIS attains strong overall extraction quality (Table~\ref{tab:sroie}).

\begin{table}[t]
\centering
\caption{\textbf{SROIE} overall and per-field accuracy.}
\label{tab:sroie}
\resizebox{0.65\linewidth}{!}{
\begin{tabular}{lcccc}
\hline \rule{0pt}{1.1\normalbaselineskip}
\textbf{Metric} & \textbf{Precision} & \textbf{Recall} & \textbf{F1} & \textbf{Field Acc.} \\
\hline \rule{0pt}{1.1\normalbaselineskip}
Overall & 0.8189 & 0.8045 & 0.8116 & 0.8045 \\
\hline \rule{0pt}{1.1\normalbaselineskip}
Field & Company & Address & Date & Total \\ \rule{0pt}{1.1\normalbaselineskip}
Accuracy & 0.8500 & 0.8500 & 0.7600 & 0.7576 \\
\hline
\end{tabular}}
\end{table}
Across synthetic stress tests and SROIE, POLARIS translates typed planning, constrained selection, and bounded repair into decision-grade outputs with explicit lineage. Compared to point OCR+rules baselines, we see higher recall under noise and precise policy/alarm signals, while preserving auditability and policy compliance by construction.

\subsection{Benchmark Implications}
\textit{The controlled synthetic suite introduced in this work provides an initial benchmark for evaluating governed agentic systems.} 
Beyond conventional precision and recall, it measures enterprise-specific dimensions such as policy compliance, audit-trace completeness, and risk-routing accuracy. 
POLARIS’s typed DAG traces, validator repair logs, and policy decisions constitute reusable evaluation primitives that enable reproducibility of compliance accuracy, routing precision, and repair effectiveness across agent frameworks. 

Future benchmarks may expand these dimensions with standardized governance metrics and multi-agent reproducibility protocols, forming a baseline schema for comparing enterprise-grade Agentic AI systems.
\section{Conclusion}
\label{sec:conclusion}
Classical back-office automation relies on brittle point tools (OCR + rules) that lack end-to-end provenance; generic agent stacks provide flexibility but often blur interfaces and policy boundaries. \textbf{POLARIS} integrates: \emph{typed DAG planning} (safe composition), \emph{diversity-constrained plan synthesis} (distinct candidates), \emph{reasoning-based selection} (multi-criterion, policy-aware choice), and a \emph{bounded repair loop} (targeted, auditable fixes). This combination yields decision-grade outputs with explicit lineage and policy justifications advancing from reactive point solutions or black-box agents to policy-aligned, valuable enterprise automation. Beyond its immediate performance gains, POLARIS introduces evaluation constructs—typed DAG lineage, governed selection logs, and validator repair traces—that can inform future benchmark design for reliable enterprise Agentic AI.

By formalizing each agentic stage as a verifiable, auditable process, POLARIS bridges the gap between research prototypes and enterprise deployment realities. It shows how policy alignment, type safety, and evaluation transparency can coexist without sacrificing adaptability or performance. The framework thus provides both a reproducible benchmark and a foundation for extending governed multi-agent reasoning to other regulated domains such as finance, supply chain, and compliance. In doing so, it contributes a concrete blueprint toward  a reliable, accountable, and operationally robust Agentic AI for the enterprise. While evaluations center on finance workflows, the design generalizes to other governed domains (supply chain, HR, compliance). Broader cross-domain validation and integration with standard AgentBench tasks remain areas of future work.

\newpage

\raggedright
\bibliographystyle{iclr2025_conference}
\bibliography{contents/references}

\newpage
\appendix
\section{Expanded Preliminaries}
\label{app:prelims_full}

\begin{table}[h]
\centering
\small
\caption{\emph{Data Processor} and \emph{Reconciliation} agents and specific roles each play in guarded execution.}
\label{tab:ops-agents-full}
\begin{tabular}{p{3.6cm}p{8.4cm}}
\hline \rule{0pt}{1.1\normalbaselineskip}
\textbf{Agent} & \textbf{Role Definitions} \\
\hline \rule{0pt}{1.1\normalbaselineskip}
\texttt{RiskControl} & Applies compliance controls and segregation-of-duties checks; assigns risk level and enumerates triggered rules. \\ \rule{0pt}{1.1\normalbaselineskip}
\texttt{RecordMatcher} & Matches invoice fields against reference ledgers or reports; flags field-level discrepancies. \\ \rule{0pt}{1.1\normalbaselineskip}
\texttt{APIAccess} & Interfaces with external ERP/Tax/Bank systems and maps responses to normalized schemas. \\ \rule{0pt}{1.1\normalbaselineskip}
\texttt{Approval} & Aggregates upstream evidence and produces a reasoned approval or rejection. \\ \rule{0pt}{1.1\normalbaselineskip}
\texttt{ReportGenerator} & Produces concise, human- and machine-readable execution summaries. \\ \rule{0pt}{1.1\normalbaselineskip}
\texttt{Scheduler} & Dependency-aware executor compiling ordered stages into a DAG for safe parallelism. \\
\hline
\end{tabular}
\end{table}

\paragraph{Task setting.}
We target document–centric workflows (e.g. invoice processing) where heterogeneous inputs such as PDFs, scans, email threads, prompts, or API events are transformed into \emph{decision-grade} outputs: validated records, approvals/rejections, anomaly alerts, and audit-ready reports. 
Let $\mathcal{X}$ denote raw inputs and $\tau\!\in\!\mathcal{T}$ a normalized task record capturing document type, metadata, and required fields.

\paragraph{Typed agents and I/O contracts.}
Each GPT-backed component (\texttt{Autogen ConversableAgent}) $A$ declares:  
(i) a capability description; (ii) typed I/O contracts with optional confidences; (iii) preconditions/postconditions; and (iv) policy and risk tags indicating eligibility, side effects, and segregation constraints.  
Agents are categorized into the following roles: Normalizer, Planning/Selection, Data Extraction, Data Processing, and Reconciliation (Table~\ref{tab:ops-agents-full}).

\paragraph{Plans as typed DAGs.}
A plan $p$ is a typed DAG $G_p=(V_p,E_p)$ whose nodes are agent invocations $(A,\iota,\sigma)$ with explicit input/output types. 
Feasibility requires type-compatible edges, compliance-order sequencing, and concurrency limits. 
The scheduler executes with completion-based semantics: a node is eligible once all its predecessors finish, enabling safe partial fan-out and aggregation of intermediate results.

\paragraph{Compliance and anomaly primitives.}
Compliance checks cover: (i) unknown vendor; (ii) threshold breach without approval; (iii) currency mismatch. 
Anomaly detection employs a vendor-wise MAD-based score
\[
z_{\mathrm{MAD}}(x;v)=\frac{|x-\mathrm{median}_v|}{1.4826\cdot\mathrm{MAD}_v},
\]
triggering when $z_{\mathrm{MAD}}>k_{\mathrm{mad}}$ (default $3.5$), with fallback to cohort/global baselines and date-sanity rules for impossible timestamps.
The routing logic for anomalies appears in Section~\ref{sec:anomaly}. 
A violation occurs if
\[
\neg E(v) \lor (E(v)\land a(i)>\tau(v)) \lor (E(v)\land c(i)\neq c^*(v)).
\]

\end{document}